\documentclass{article}

\usepackage{verbatim}
\usepackage[final,nonatbib]{nips_2016}
\usepackage{graphicx}
\usepackage{subfigure}
\usepackage[utf8]{inputenc} % allow utf-8 input
\usepackage[T1]{fontenc}    % use 8-bit T1 fonts
\usepackage{url}            % simple URL typesetting
\usepackage{booktabs}       % professional-quality tables
\usepackage{amsfonts}       % blackboard math symbols
\usepackage{nicefrac}       % compact symbols for 1/2, etc.
\usepackage{microtype}      % microtypography

\title{Neuro-symbolic EDA-based Optimisation using ILP-enhanced DBNs}

\author{
  Sarmimala Saikia\\ TCS Research, New Delhi \\sarmimala.saikia@tcs.com \And
  Lovekesh Vig\\ TCS Research, New Delhi \\ lovekesh.vig@tcs.com \And 
  Ashwin Srinivasan \\ Department of Computer Science \\BITS Goa \\ ashwin@goa.bits-pilani.ac.in \And
  Gautam Shroff \\ TCS Research, New Delhi \\ gautam.shroff@tcs.com\And 
  Puneet Agarwal \\ TCS Research, New Delhi \\ puneet.a@tcs.com \And 
  Richa Rawat \\ TCS Research, New Delhi \\ rawat.richa@tcs.com   
}

\begin{document}
% \nipsfincalcopy is no longer used

\maketitle

\begin{abstract}
We investigate solving discrete optimisation problems using the ‘estimation of distribution’ (EDA) approach via a novel combination of deep belief networks
(DBN) and inductive logic programming (ILP). While DBNs are used to learn the structure of successively ‘better’ feasible solutions, ILP enables the 
incorporation of domain-based background knowledge related to the goodness of solutions. Recent work showed that ILP could be an effective way to use 
domain knowledge in an EDA scenario. However, in a purely ILP-based EDA, sampling successive populations is either inefficient or not straightforward. 
In our Neuro-symbolic EDA, an ILP engine is used to construct a model for good solutions using domain-based background knowledge. 
These rules are introduced as Boolean features in the last hidden layer of DBNs used for EDA-based optimization. 
This incorporation of logical ILP features requires some changes while training and sampling from DBNs: (a) our DBNs need to be trained with data for 
units at the input layer as well as some units in an otherwise hidden layer; and (b) we would like the samples generated to be drawn from instances entailed by 
the logical model. We demonstrate  the viability of our approach on instances of two optimisation problems: predicting optimal depth-of-win for the KRK endgame, 
and job-shop scheduling. Our results are promising: (i) On each iteration of distribution estimation, samples obtained with an ILP-assisted DBN 
have a substantially greater proportion of good solutions than samples generated using a DBN without ILP features; and (ii) On termination of 
distribution estimation, samples obtained using an ILP-assisted DBN contain more near-optimal samples than samples from a DBN without ILP features. 
Taken together, these results suggest that the use of ILP-constructed theories could be useful for incorporating complex domain-knowledge into deep 
models for estimation of distribution based procedures. 
\end{abstract}

\section{Introduction}
There are many real-world planning problems for which domain knowledge is
qualitative, and not easily encoded in a form suitable for numerical optimisation.
Here, for instance, are some guiding principles that are followed by the Australian Rail
Track Corporation when scheduling trains: (1) If a healthy
Train is running late, it should be given equal preference to other healthy Trains;
(2) A higher priority train should be given preference to a lower priority train,
provided the delay to the lower priority train is kept to a minimum; and so on. It
is evident from this that train-scheduling may benefit from knowing if a train is
healthy, what a trains priority is, and so on. But are priorities and train-health
fixed, irrespective of the context? What values constitute acceptable delays to a
low-priority train? Generating good train-schedules will require a combination
of quantitative knowledge of train running times and qualitative knowledge
about the train in isolation, and in relation to other trains. In this paper, we
propose a heuristic search method, that comes under the broad category of an
estimation distribution algorithm (EDA). EDAs iteratively generate better solutions
for the optimisation problem using machine-constructed models. Usually
EDAs have used generative probabilistic models, such as Bayesian Networks,
where domain-knowledge needs to be translated into prior distributions and/or
network topology. In this paper, we are concerned with problems for which such a
translation is not evident. Our interest in ILP is that it presents perhaps one of
the most flexible ways to use domain-knowledge when constructing models. Recent work 
has shown that ILP models incorporating background knowledge were able to generate better quality solutions in each EDA iteration \cite{ash16}.
However, efficient sampling is not straightforward and ILP is unable to utilize the discovery of 
high level features as efficiently as deep generative models. 

While neural models have been used for optimization \cite{vinyals2015pointer}, in this paper we attempt to combine the sampling and feature discovery power of deep generative models with the 
background knowledge captured by ILP for optimization problems that require domain knowledge. The rule based features discovered by the ILP engine are appended to the higher layers
of a Deep Belief Network(DBN) while training. A subset of the features are then clamped on while sampling to generate 
samples consistent with the rules. This results in consistently improved sampling which has a cascading positive effect on successive iterations of EDA based optimization procedure. 
The rest of the paper is organised as follows. Section 2 provides a brief description
of the EDA method we use for optimisation problems. Section 2.1
describes how ILP can be used within the iterative loop of an EDA for discovering rules that would distinguish good samples from bad. 
Section 3 Describes how RBMs can be used to generate samples that conform to the rules discovered by the ILP engine.
Section 4 describes an empirical evaluation demonstrating the improvement in the discovery of optimal solutions,  followed by conclusions in Section 5.
\section{EDA for optimization}
\label{sec:eda-for-optimization}
The basic EDA approach we use is the one proposed by the MIMIC algorithm \cite{mimic}. Assuming
that we are looking to minimise an objective function $F(\textbf{x})$, where $\textbf{x}$ is
an instance from some instance-space ${\cal X}$, the approach first constructs an appropriate
machine-learning model to discriminate between samples of lower and higher value, i.e., $F(\mathbf{x}) \leq \theta$ and
$F(\mathbf{x}) > \theta$, and then generates samples using this model
% to generate a population for the next iteration,
% while also lowering $\theta$. This is described by the procedure in Fig.~\ref{fig:eoms}.
\begin{figure}[htb]
{\small{
\begin{description}
\item[Procedure EODS:] Evolutionary Optimisation using DBNs for Sampling
\begin{enumerate}
    \item Initialize population $P:=\{\textbf{x}_i\}$; $\theta:=\theta_0$
    \item while not converged do
        \begin{enumerate}
            \item for all $\mathbf{x}_i$ in $P$ $label(\textbf{x}_i)$ := 1 if
            $F(\textbf{x}_i) \leq \theta$ else $label(\textbf{x}_i)$ := 0
            \item train DBN $M$ to discriminate between 1 and 0 labels
                    i.e.,
                    $P(\textbf{x}: label(\textbf{x}) = 1 | M ) > P(\textbf{x}: label(\textbf{x}) = 0 | M )$
            \item regenerate $P$ by repeated sampling using model $M$
            \item reduce threshold $\theta$
        \end{enumerate}
\item return $P$
\end{enumerate}
\end{description}
}}
\caption{Evolutionary optimisation using a network model to generate samples.}
\label{fig:eoms}
\end{figure}

Here we use Deep Belief Networks (DBNs) \cite{hinton2011} for modeling our data distribution, and for generating
samples for each iteration of MIMIC. Deep Belief Nets (DBNs) are generative models that are composed of
multiple latent variable models called Restricted Boltzman Machines (RBMs).
In particular, as part of our larger optimization algorithm, we wish to repeatedly
train and then sample from the trained DBN in order to reinitialize our sample population for the next
iteration as outlined in Figure \ref{fig:eoms}. In order to accomplish this, while training we append a
single binary unit (variable) to the highest hidden layer of the DBN, and assign it a value $1$
when the value of the sample is below $\theta$ and a value $0$ if the value is above $\theta$.
During training that this variable, which we refer to as the separator variable,
learns to discriminate between good and bad samples. To sample from the DBN  we additionally clamp our separator variable
to $1$ so as to bias the network to produce good samples, and preserve the DBN weights
from the previous MIMIC iteration to be used as the initial weights for the
subsequent iteration. This prevents retraining on the same data repeatedly as
the training data for one iteration subsumes the samples from the previous iteration.  

We now look at how ILP models can assist DBNs constructed for this purpose.
\section{EDA using ILP-assisted DBNs}
\label{sec:BK}
\subsection{ILP}
\label{sec:ilp}
The field of Inductive Logic Programming (ILP) has made steady
progress over the past two and half decades, in advancing the theory,
implementation and application of logic-based relational learning.
A characteristic of this form of machine-learning is that
data, domain knowledge and models are usually---but not always---expressed in a
subset of first-order logic, namely logic programs.
Side-stepping for the moment the question ``why logic programs?'',
domain knowledge (called {\em background
knowledge\/} in the ILP literature) can be encodings of heuristics, rules-of-thumb,
constraints, text-book knowledge and so on. It is evident
that the use of some variant of first-order logic
enable the automatic construction of models that use relations
(used here in the formal sense of a truth value assignment to $n$-tuples).
Our interest here is in a form of relational learning
concerned with the identification of functions
(again used formally, in the sense of being a uniquely
defined relation) whose domain is the set of instances in the data.
An example is the construction of new {\em features\/} for data analysis
based on existing relations (``$f(m) = 1$ if a molecule $m$ has 3 or more
benzene rings fused together otherwise $f(m) = 0$'': here concepts
like benzene rings and connectivity of rings are generic relations provided
in background knowledge). 

There is now a growing body of research that suggests that
ILP-constructed relational features can substantially improve the predictive power of a
statistical model (see, for example: \cite{JoshiRS08,Amrita12,Specia_09,RamakrishnanJBS07,SpeciaSRN06}).
Most of this work has concerned itself with discriminatory models, although there have been cases
where they have been incorporated within generative models.
In this paper, we are interested in their use within a deep network model used for generating
samples in an EDA for optimisation in Procedure EODS in Fig.~\ref{fig:eoms}.
\subsection{ILP-assisted DBNs}
\label{sec:dbnilp}
Given some data instances
$\mathbf{x}$ drawn from a set of instances ${\cal X}$ and domain-specific background knowledge,
let us assume the ILP engine will be used to construct a model for discriminating between
two classes (for simplicity, called $good$ and $bad$). The ILP engine constructs a model for $good$ instances using rules of the form
$h_j:$ $Class(\mathbf{x},good) \leftarrow {Cp}_j(\mathbf{x})$.\footnote{We note that
in general for ILP $\mathbf{x}$ need not be
restricted to a single object and can consist of arbitrary tuples of
objects and rules constructed by the ILP engine would more generally
be
$h_j:~Class(\langle {\mathbf x_1},{\mathbf x_2},\ldots,{\mathbf x_n}\rangle,c)$
$\leftarrow {Cp}_j(\langle \mathbf{x_1},{\mathbf x_2},\ldots,{\mathbf x_n}\rangle)$. But we
do not require rules of this kind here.}
 ${Cp}_j: {\cal X}
\mapsto \{0,1\}$ denotes a ``context predicate''. A context predicate
corresponds to a conjunction of literals that evaluates to $TRUE$ (1) or
$FALSE$ (0) for any element of ${\cal X}$.
For meaningful features we will usually require that a ${Cp}_j$ contain
at least one literal; in logical terms, we therefore require the
corresponding $h_j$ to be definite clauses with at least two literals.
A rule $h_j: Class({\mathbf x},good) \leftarrow
{Cp}_j({\mathbf x})$, is converted to a feature $f_j$ using a one-to-one mapping as
follows: $f_j({\mathbf x}) = 1~{\mathrm{iff}}~{Cp}_j({\mathbf x}) = 1$ (and $0$ otherwise).
We will denote this function as $Feature$. Thus $Feature(h_j) = f_j$,
${Feature}^{-1}(f_j) = h_j$. We will also sometimes refer to
$Features(H) = \{f: h \in H ~\mathrm{and}~ f = Feature(h)\}$ and
$Rules(F) = \{h: f \in F ~\mathrm{and}~ h = {Features}^{-1}(f)\}$.

\begin{figure*}[ht!]
\centering      
  \subfigure[]{%
        \begin{minipage}[h]{0.48\textwidth}
            \centering
       \includegraphics*[scale=.18]{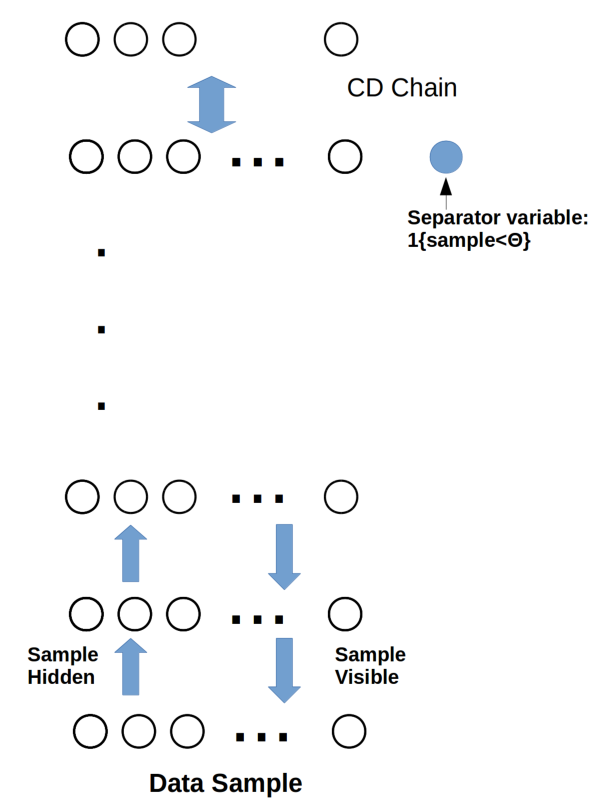}%.eps}
        \end{minipage}
        }
  \subfigure[]{%
        \begin{minipage}[h]{0.48\textwidth}
            \centering
            \includegraphics*[scale=.18]{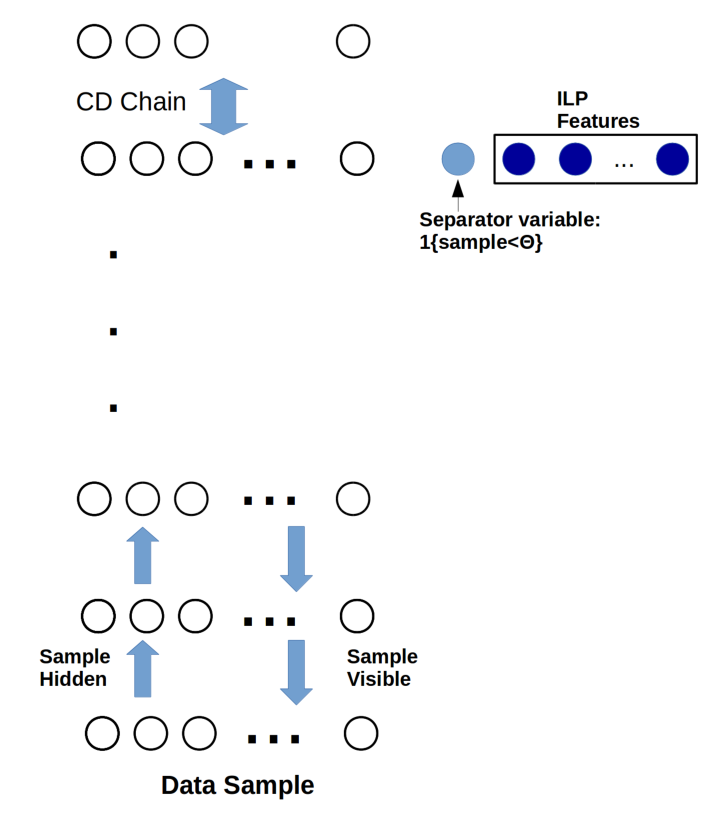}%.eps}
        \end{minipage}
        }
    \caption{ Sampling from a DBN (a) with just a separator variable (b) with ILP features}
\label{fig:train-dbn}
\end{figure*}

Each rule in an ILP model is thus converted to a single Boolean feature, and the model
will result in a set of Boolean features. Turning now to the EODS procedure
in Fig.~\ref{fig:eoms}, we will construct ILP models for discriminating between
$F(\textbf{x}) \leq \theta)$ ($good$) and $F(\textbf{x}) > \theta$ ($bad$).
Conceptually, we treat the ILP-features as high-level features for a deep belief network,
and we append the data layer of the highest level RBM with the values of
the ILP-features for each sample as shown in Fig \ref{fig:train-dbn}.
\subsection{Sampling from the Logical Model}
Recent work \cite{ash16} suggests that if samples can be drawn from the success-set of the
ILP-constructed model\footnote{These are the instances entailed by the model along with the background
knowledge, which---assuming the rules are not recursive---we take to be the union of the success-sets
of the individual rules in the model.} then the efficiency of identifying near-optimal solutions could
be significantly enhanced. A straightforward approach of achieving this with an ILP-assisted
DBN would appear to be to clamp all the ILP-features, since this would bias the
the samples from the network to sample from the intersection of the success-sets of the corresponding
rules (it is evident that instances in the intersection are guaranteed to be in the success-set sought).
However this will end up being unduly restrictive, since samples sought are not ones that satisfy all
rules, but {\em at least\/} one rule in the model. The obvious modification would be to clamp subsets
of features. But not all samples from a subset of features may be appropriate.

With a subset of features clamped, there is an additional complication that arises due
to the stochastic nature of the DBN's hidden units. This makes it possible for the DBN's unit
corresponding to a logical feature $f_j$ to have the value $1$ for an instance ${\mathbf x}_i$,
but for ${\mathbf{x_i}}$ not to be entailed by the background knowledge and logical rule $h_j$.

In turn, this means that for a feature-subset with values clamped,
samples may be generated from outside the success-set of corresponding rules involved.
Given background knowledge $B$, we say a
sample instance ${\mathbf x}$ generated by clamping a set of features $F$ is
{\em aligned\/} to $H = Rules(F)$, iff  
$B \wedge H \models {\mathbf x}$ (that is, ${\mathbf x}$ is entailed by $B$ and $H$).

A procedure to bias sampling of instances from the success-set of the
logical model constructed by ILP is shown in Fig.~\ref{fig:align}.
\begin{figure*}[htb]
\begin{description}
\item[Given:] Background knowledge $B$; a set of rules $H =  \{h_1,h_2,\ldots,h_N\}$l; 
    a DBN with $F = \{f_1, f_2, \ldots, f_N\}$ as high-level features ($f_i = Feature(h_i)$);
    and a sample size $M$
\item[Return:] A set of samples $\{{\mathbf x}_1, {\mathbf x}_2, \ldots, {\mathbf x}_M\}$
        drawn from the success-set of $B \wedge H$.  
\end{description} 
\begin{enumerate}
    \item $S:= \emptyset, k=0$
    \item while $|k| \leq N$ do
        \begin{enumerate}
             \label{step:randselect}
            \item Randomly select a subset $F_k$ of size $k$ features from $F$
            \item Generate a small sample set $X$ clamping features in $F_k$
            	\item for each sample in $x \in X$ and each rule $h_j$, set $count_k=0$
            	\begin{enumerate}	 
             	\item if $x \in$ success-set where $(f_j(x) =1) => (x \in$ success-set($B$ and $h_j))$ 	
             	
					 $count_k=count_k+1$             	
            	\end{enumerate}
            	
            %\item if $B \wedge Rules(F_k) \models {\mathbf x}$ then $S:= S \union \{{\mathbf x}\}$
        \end{enumerate}
    \item Generate $S$  by clamping $k$ features where  $count_k=max(count_1,count_2...count_N)$
  \item return $S$
\end{enumerate}
\caption{A procedure to generate samples aligned to a logical model $H$ constructed by
    an ILP engine.}
\label{fig:align}
\end{figure*}
\section{Empirical Evaluation}
\label{sec:expt}
\subsection{Aims}
\label{sec:aim}
Our aims in the empirical evaluation are to investigate the
following conjectures:
\begin{enumerate}
\item[(1)] On each iteration, the EODS procedure will yield
        better samples with ILP features than without
\item[(2)] On termination, the EODS procedure will yield more
        near-optimal instances with ILP features than without.
\item[(3)] Both procedures do better than random sampling from the initial training set.
%\item[(3)] Increasing relevant domain-specific background knowledge provided to the
        %ILP engine results in better samples from the EODS procedure.
\end{enumerate}

\noindent
It is relevant here to clarify what the comparisons are intended in
the statements above. Conjecture (1) is essentially a statement
about the gain in precision obtained by using ILP features. Let us denote
$Pr(F(\textbf{x}) \leq \theta)$ the probability of generating an instance $\mathbf{x}$ with cost at
most $\theta$ without ILP features to guide sampling, and
by $Pr(F(\textbf{x}) \leq \theta|M_{k,B})$ the probability of obtaining such an instance with
 ILP features $M_{k,B}$ obtained on iteration $k$ of the EODS
procedure using some domain-knowledge $B$.
%That is,
%$Pr((F(\textbf{x}) \leq \theta|\textbf{x} \in {\cal X} ~\mathrm{and}~B \wedge M_{k,B} \models \textbf{x}\}$
(note if $M_{k,B} = \emptyset$, then we will mean
$Pr(F(\textbf{x}) \leq \theta|M_{k,B})$ = $Pr(F(\textbf{x}) \leq \theta)$).
Then for (1) to hold, we would require
$Pr(F(\textbf{x}) \leq \theta_k | M_{k,B}) > Pr(F(\textbf{x}) \leq \theta_k)$.
given some relevant $B$. We will estimate the probability on the lhs from
the sample generated using the model, and the probability on the rhs from
the datasets provided.
Conjecture (2) is related to the gain in recall obtained by using the model,
although it is more practical to examine actual numbers of near-optimal instances
(true-positives in the usual terminology). We will compare the numbers of near-optimal
in the sample generated by the DBN model with ILP features, to those obtained
using the DBN alone.
%For (2), let $P_k$ denotes the instances generated on iteration $k$ of EODS.
%Then the probability of obtain a near-optimal instance in iteration $k$ is
%by $Pr(\textbf{x} ~\in~ P_k | F(\textbf{x}) \leq \theta^*)$,
%and the corresponding probability using simple random sampling is
%$Pr(F(\textbf{x}) \leq \theta^*)$
%Thus, for (2) to hold,
%$Pr(\textbf{x} ~\in~ P_k | F(\textbf{x}) \leq \theta^*)$
%$>$
%$Pr(F(\textbf{x}) \leq \theta^*)$.
%%Finally, (3) requires
%%$Pr(F(\textbf{x}) \leq \theta_k| M_{k,(B_1 \cup B_2)}) > Pr(F(\textbf{x}) \leq \theta_k| M_{k,B_1})$
%% given relevant $B_1$ and $B_2$.
%The empirical evalution examines
%these comparisons using empirical estimates of the probabilities.
\subsection{Materials}
\label{sec:materials}
\subsubsection{Data}
We use two synthetic datasets, one arising from the KRK chess endgame (an endgame with just White King, White Rook and Black King
on the board), and the other a restricted, but nevertheless hard $5 \times 5$ job-shop scheduling (scheduling 5 jobs taking varying lengths of
time onto 5 machines, each capable of processing just one task at a time).

The optimisation problem we examine for the KRK endgame is to predict the depth-of-win with optimal play \cite{bain:krkwin}.
Although aspect of the endgame has not been as popular in ILP as task of predicting ``White-to-move position is illegal''
\cite{bain:krkillegal}, it 
offers a number of advantages as a {\em Drosophila\/} for optimisation problems of the kind we are interested.
First, as with other chess endgames, KRK-win is a complex, enumerable domain for which
there is complete, noise-free data. Second, optimal ``costs'' are known for all data instances.
Third, the problem has been studied by
chess-experts at least since Torres y Quevado built a machine, in 1910, capable of playing the KRK endgame. This
has resulted in a substantial amount of domain-specific knowledge. We direct the reader to \cite{breda:thesis}
for the history of automated methods for the KRK-endgame. For us, it suffices to treat the problem as a form
of optimisation, with the cost being the depth-of-win with Black-to-move, assuming minimax-optimal play.
In principle, there are ${64}^3 ~\approx 260,000$ possible positions for the KRK endgame,
not all legal. Removing illegal
positions, and redundancies arising from symmetries of the board reduces the size of the
instance space to about $28,000$ and the distribution shown in
Fig.~\ref{fig:distr}(a). The sampling task here is to generate instances with depth-of-win equal to $0$.
Simple random sampling has a probability of about $1/1000$ of generating such an instance once redundancies
are removed.

The job-shop scheduling problem is less controlled than the chess endgame, but is nevertheless representative
of many real-life applications (like scheduling trains), and in general, is known to be computationally
hard.
\begin{figure}
\begin{minipage}[h]{0.40\textwidth}
\centering
{\scriptsize{
\begin{tabular}{|cl|cl|} \hline
Cost & Instances   & Cost & Instances \\ \hline
0     & 27 (0.001) & 9     & 1712 (0.196) \\
1     & 78 (0.004)  & 10     &1985 (0.267) \\
2     & 246 (0.012) & 11    & 2854 (0.368) \\
3     & 81  (0.152) & 12    & 3597 (0.497) \\
4     & 198 (0.022) & 13    & 4194 (0.646) \\
5     & 471 (0.039) & 14    & 4553 (0.808)\\
6     & 592 (0.060) & 15    &  2166 (0.886) \\
7     & 683 (0.084) & 16    & 390 (0.899)  \\
8     & 1433 (0.136)   & draw  & 2796 (1.0) \\ \hline 
\multicolumn{4}{l}{Total Instances: 28056} \\
\end{tabular}
}}
\begin{center}
(a) Chess
\end{center}
\end{minipage}
\begin{minipage}[h]{0.60\textwidth}
\vspace*{1cm}
\centering
{\scriptsize{
\begin{tabular}{|cl|cl|} \hline
Cost & Instances & Cost & Instances \\ \hline
400--500 & 10 (0.0001) & 1000--1100   & 24067 (0.748) \\
500--600 & 294 (0.003) & 1100--1200  &  15913 (0.907) \\
600--700 & 2186 (0.025)  & 1200--1300  & 7025 (0.978) \\
700--800 & 7744 (0.102) & 1300--1400  &  1818 (0.996) \\
800--900 & 16398 (0.266) & 1400--1500 & 345 (0.999) \\
900--1000& 24135 (0.508) & 1500--1700 & 66  (1.0) \\ \hline 
\multicolumn{4}{l}{Total Instances: 100000} \\
\end{tabular}
}}
\begin{center}
(b) Job-Shop
\end{center}
\end{minipage}
\caption{Distribution of cost values. The number in parentheses are
    cumulative frequencies.}
\label{fig:distr}
\end{figure}

\noindent
Data instances for Chess are in the form of 6-tuples, representing the rank and file (X and Y values) of the 3 pieces
involved. For the RBM, these are encoded as 48 dimensional binary vector where every eight bits represents a one hot encoding of the pieces' rank or file.
 At each iteration $k$ of the EODS procedure, some instances with depth-of-win $\leq \theta_k$ and the rest
with depth-of-win $> \theta_k$ are used to construct the ILP model, and the resulting features are appended to train the RBM model as described in Section \ref{sec:dbnilp}.\footnote{The $\theta_k$ values are pre-computed
assuming optimum play. We note that when constructing a model on iteration $k$, it is permissible to use
all instances used on iterations $1,2,\ldots,(k-1)$ to obtain data for model-construction.}

Data instances for Job-Shop are in the form of schedules, with associated start- and end-times for each task on
a machine, along with the total cost of the schedule. On iteration $i$ of the EODS procedure,
models are to be constructed to predict if the cost of schedule will be $\leq \theta_i$ or otherwise.\footnote{The
total cost of a schedule includes any idle-time, since for each job, a task
before the next one can be started for that job. Again, on iteration $i$,
it is permissible to use data from previous iterations.}
\subsubsection{Background Knowledge}
%For each problem, we examine two subsets of background definitions, which we
%denote by $B_{low}$ and $B_{high}$. $B_{low}$ consists of background predicates with low relevance
%to the problem. $B_{high}$ adds predicates of higher relevance to those in $B_{low}$.
%For Chess, $B_{low}$ simply consists of predicates related to the geometry of the board, namely
%a comparison of rank and files (X and Y values), and square-adjacency. These are identical
%to the predicates used in prior work on the endgame \cite{bain:krk}. For $B_{high}$, we extend

For Chess, background predicates encode the following (WK denotes the White King, WR the White Rook, and BK the Black King):
(a) Distance between pieces WK-BK, WK-BK, WK-WR;
(b) File and distance patterns: WR-BK, WK-WR, WK-BK;
(c) ``Alignment distance'': WR-BK;
(d) Adjacency patterns: WK-WR, WK-BK, WR-BK;
(e) ``Between'' patterns: WR between WK and BK, WK between WR and BK, BK
    between WK and WR;
(f) Distance to closest edge: BK;
(g) Distance to closest corner: BK;
(h) Distance to centre: WK; and
(i) Inter-piece patterns: Kings in opposition, Kings almost-in-opposition,
    L-shaped pattern.
We direct the reader to \cite{breda:thesis} for the history of using these concepts, and their definitions. A sample rule generated for Depth<=2 is that the distance between the files of the two kings be greater than or equal to zero, and that the  ranks of the kings are seperated bya a distance of less than five and those of the white king and the rook by less than 3.

%For Job-Shop, $B_{low}$ consists simply of task-ordering in a schedule (task $1$ of job $J_i$ occurs before
%task $1$ of job $J_k$, {\em etc\/.}). $B_{high}$ consists of $B_{low}$ and, in addition, 
For Job-Shop, background predicates encode:
(a) schedule job $J$ ``early'' on machine $M$ (early means first or second);
(b) schedule job $J$ ``late'' on machine $M$ (late means last or second-last);
(c) job $J$ has the fastest task for machine $M$;
(d) job $J$ has the slowest task for machine $M$;
(e) job $J$ has a fast task for machine $M$ (fast means the fastest or second-fastest);
(f) Job $J$ has a slow task for machine $M$ (slow means slowest or second-slowest);
(g) Waiting time for machine $M$;
(h) Total waiting time;
(i) Time taken before executing a task on a machine. Correctly, the predicates for (g)--(i)
encode upper and lower bounds on times, using the standard inequality predicates $\leq$ and $\geq$.
\subsubsection{Algorithms and Machines}
The ILP-engine we use is Aleph (Version 6, available from A.S. on request). 
All ILP theories were constructed on an Intel Core i7 laptop computer, using
VMware virtual machine running Fedora 13, with an allocation of 2GB for
the virtual machine. The Prolog compiler used was Yap,
version 6.1.3\footnote{\url{http://www.dcc.fc.up.pt/~vsc/Yap/}}. The RBM was implemented in the Theano library, and run on an NVidia Tesla K-40 GPU Card. 
\subsection{Method}
\label{sec:method}
Our method is straightforward:

\begin{itemize}
\item[] For each optimisation problem, and domain-knowledge $B$:
    \begin{itemize}
        \item[] Using a sequence of threshold values
                $\langle \theta_1, \theta_2, \ldots, \theta_n \rangle$
        on iteration $k$ ($1 \leq k \leq n$) for the EODS procedure:
        \begin{enumerate}
            \item Obtain an estimate of $Pr(F(\textbf{x}) \leq \theta_k)$ using
                                a DBN with a separator variable;
            \item Obtain an estimate of $Pr(F(\textbf{x}) \leq \theta_k|M_{k,B})$
                                    by constructing an ILP model for
                                    discriminating between $F(\textbf{x}) \leq \theta_k$ and
                                    $F(\textbf{x}) > \theta_k$. Use the features learnt by the ILP model to guide the DBN sampling.
            \item Compute the ratio of
                                $Pr(F(\textbf{x}) \leq \theta_k|M_{k,B})$
                                to $P(F(\textbf{x}) \leq \theta_k)$
         \end{enumerate}
    \end{itemize}
\end{itemize}

\noindent
The following details are relevant:
\begin{itemize}
\item The sequence of thresholds for Chess are $\langle 8, 4, 2, 0 \rangle$. For
    Job-Shop, this sequence is $\langle 900, 890, 880...600 \rangle$; Thus,
    $\theta^*$ = 0 for Chess and $600$ for Job-Shop, which means we
    require exactly optimal solutions for Chess.
\item Experience with the use of ILP engine used here (Aleph)
    suggests that the most sensitive parameter is the one defining a lower-bound on the 
        precision of acceptable clauses (the $minacc$ setting in Aleph).
        We report experimental results obtained with $minacc=0.7$, which has
        been used in previous experiments with the KRK dataset.
        The background knowledge for Job-Shop does not appear to
        be sufficiently powerful to allow the identification of
        good theories with short clauses. That is, the usual Aleph
        setting of upto 4 literals per clause leaves most of the
        training data ungeneralised. We therefore allow an
        upper-bound of upto 10 literals for Job-Shop, with
        a corresponding increase in the number of search nodes to 10000
        (Chess uses the default setting of 4 and 5000 for these parameters).
\item In the EODS procedure, the initial sample is obtained using a uniform
    distribution over all instances. Let us call this
        $P_0$. On the first iteration of EODS ($k=1$), the datasets
        ${E_1}^+$ and ${E_1}^-$ are obtained by computing the (actual) costs for
        instances in $P_0$, and an ILP model $M_{1,B}$, or simply $M_1$, constructed.
    A DBN model is constructed both with and without ILP features.
        We obtained samples from the DBN with $CD_6$ or by running the Gibbs chain for six iterations. 
         On each iteration $k$, an estimate of $Pr(F(\textbf{x}) \leq \theta_k)$ can be
        obtained from the empirical frequency distribution of instances with values
        $\leq \theta_k$ and $> \theta_k$. For the synthetic problems here,
        these estimates are in Fig.~\ref{fig:distr}.
        For $Pr(F(\textbf{x}) \leq \theta_k|M_{k,B})$, we use
        obtain the frequency of $F(\textbf{x}) \leq \theta_k$ in $P_k$
\item Readers will recognise that the ratio of
        $Pr(F(\textbf{x}) \leq \theta_k|M_{k,B})$
        to $P(F(\textbf{x}) \leq \theta_k)$
    is equivalent to computing the gain in precision obtained
    by using an ILP model over a non-ILP model. Specifically, if this ratio
    is approximately $1$, then there is no value in using the ILP model. The
    probabilities computed also provide one way of estimating sampling efficiency
    of the models (the higher the probability, the fewer samples will be needed
    to obtain an instance $\mathbf{x}$ with $F(\mathbf{x}) \leq \theta_k$).
\end{itemize}
\subsection{Results}
\label{sec:results}
\begin{figure*}[ht!]
\centering      
  \subfigure[]{%
        \begin{minipage}[h]{0.48\textwidth}
            \centering
       \includegraphics*[width=6cm]{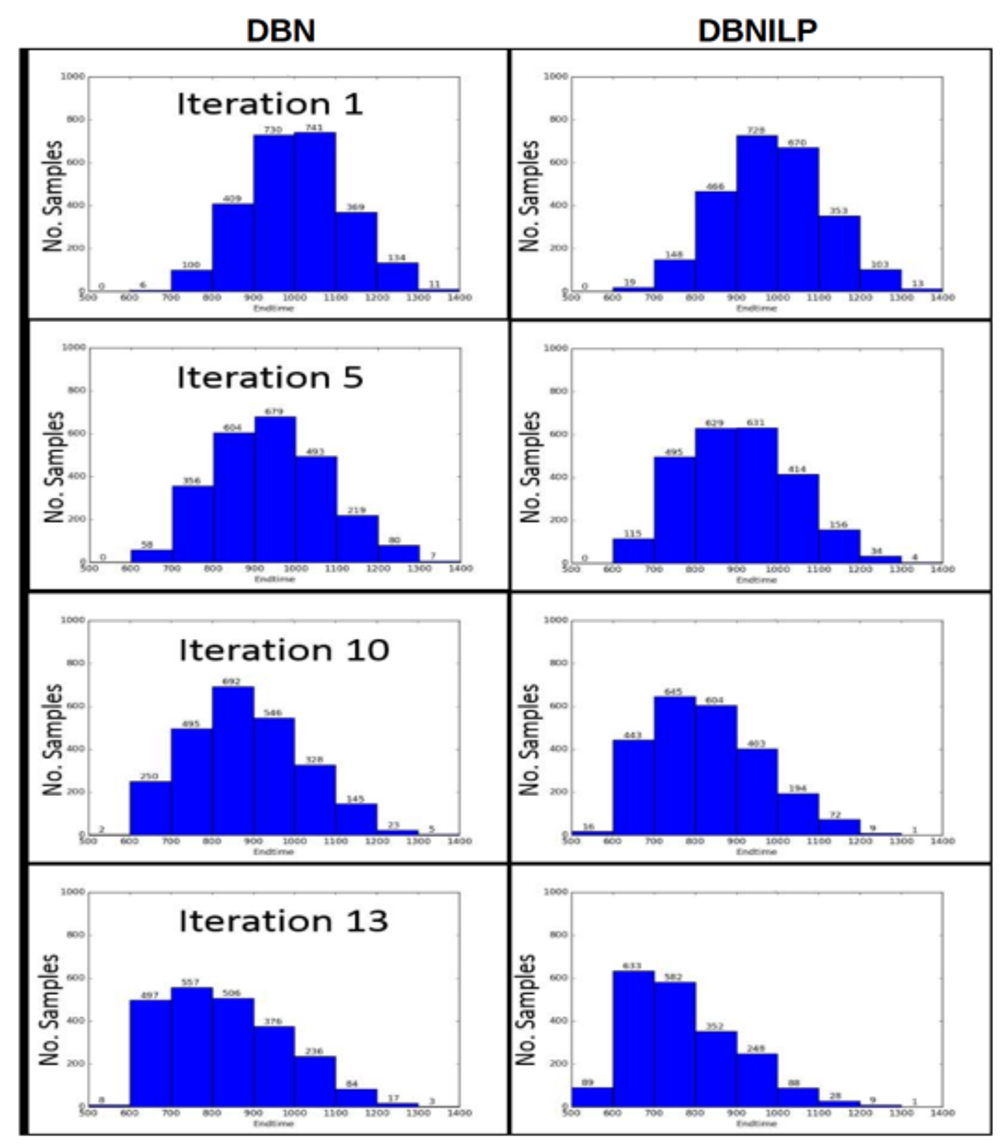}%.eps}
        \end{minipage}
        }
  \subfigure[]{%
        \begin{minipage}[h]{0.48\textwidth}
            \centering
            \includegraphics*[scale=.21]{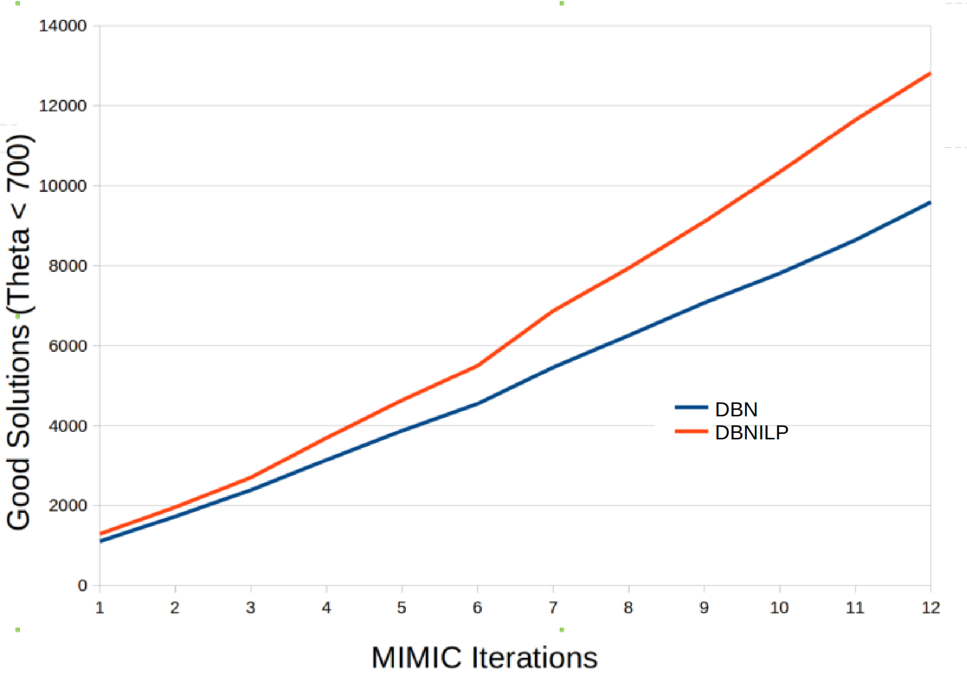}%.eps}
        \end{minipage}
        }
    \caption{Impact of ILP on EODS procedure for Job Shop (a) Distribution of  solution Endtimes generated on iterations 1, 5, 10 and 13 with and without ILP (b) Cumulative semi-optimal solutions obtained with and without ILP features over 13 iterations}
\label{fig:cascade}
\end{figure*}

Results relevant to conjectures (1) and (2) are tabulated in
Fig.~\ref{fig:results} and Fig.~\ref{fig:results2}.
The principal conclusions that can drawn from the results are these:
\begin{enumerate}
\item[(1)] For both problems, and every threshold value $\theta_k$,
    the probabilty of obtaining instances
        with cost at most $\theta_k$ with ILP-guided RBM sampling is
        substantially higher than without ILP. This provides evidence
        that ILP-guided DBN sampling results in better samples than
        DBN sampling alone(Conjecture 1);
\item[(2)] For both problems and every threshold value $\theta_k$, 
        samples obtained with ILP-guided sampling contain a 
        substantially higher number of near-optimal instances
        than samples obtained using a DBN alone (Conjecture 2)
\end{enumerate}
\noindent 
Additionally, Fig.~\ref{fig:cascade} demonstrates the cumulative impact of ILP on (a) the distribution of good solutions obtained and (b)the cascading improvement over the DBN alone for the Job Shop problem. The DBN with ILP was able to arrive at the optimal solution within 10 iterations.  
\begin{figure}
\begin{minipage}[h]{0.5\textwidth}
\centering
\begin{tabular}{|l|c|c|c|c|}
\hline
 Model& \multicolumn{4}{|c|}{$Pr(F(\mathbf{x}) \leq \theta_k|M_k)$} \\ \cline{2-5}
       & $k=1$      & $k=2$    & $k=3$     &$k=4$ \\ \hline
None   & $0.134$    & $0.042$  & $0.0008$   &$0.0005$  \\[6pt]
DBN    & $0.220$     & $0.050$   & $0.015$   &$0.0008$ \\ [6pt]
DBNILP & $0.345$    & $0.111$  & $0.101$   &$0.0016$ \\ \hline 
\end{tabular}
\begin{center}
(a) Chess
\end{center}
\end{minipage}
\begin{minipage}[h]{0.5\textwidth}
% \vspace*{0.3cm}
\centering
\begin{tabular}{|l|c|c|c|c|}
\hline
 Model & \multicolumn{4}{|c|}{$Pr(F(\mathbf{x}) \leq \theta_k|M_k)$} \\ \cline{2-5}
       & $k=1$ & $k=2$ & $k=3$ & $k=4$\\ \hline
None  & $0.040$ & $0.036$ & $0.029$& $0.024$\\ [6pt]
DBN   & $0.209$ & $0.234$ & $0.248$ & $0.264$ \\ [6pt]
DBNILP & $0.256$   & $0.259$  & $0.268$& $0.296$ \\ \hline 
\end{tabular}
\begin{center}
(b) Job-Shop
\end{center}
\end{minipage}
\caption{Probabilities of obtaining good instances $\mathbf{x}$ for each iteration
    $k$ of the EODS procedure. That is, the column $k=1$ denotes
    $P(F(\mathbf{x}) \leq \theta_1$ after iteration $1$; the column $k=2$ denotes
    $P(F(\mathbf{x}) \leq \theta_2$ after iteration $2$ and so on.
        In effect, this is an estimate of the precision when predicting
        $F(\mathbf{x}) \leq \theta_k$.
    ``None'' in the model column stands for
    probabilities of the instances, corresponding to simple random sampling
        ($M_k = \emptyset$).
     }
\label{fig:results}

\end{figure}

\begin{figure}
\begin{minipage}[h]{0.5\textwidth}
\centering
\begin{tabular}{|l|c|c|c|c|} \hline
Model & \multicolumn{4}{|c|}{Near-Optimal Instances} \\ \cline{2-5}
      & $k=1$ & $k=2$ & $k=3$ &$k=4$ \\ \hline
DBN  & 5/27   & 11/27 & 11/27 & 12/27 \\[6pt]
DBNILP   &  3/27 &  17/27 & 21/27 & 22/27 \\ \hline
\end{tabular}
\begin{center}
(a) Chess
\end{center}
\end{minipage}
\begin{minipage}[h]{0.5\textwidth}
% \vspace*{0.6cm}
\centering
\begin{tabular}{|l|c|c|c|c|} \hline
Model & \multicolumn{3}{|c|}{Near-Optimal Instances} \\ \cline{2-4}
      & $k=11$ & $k=12$ & $k=13$ \\ \hline
DBN  & 7/304 & 10/304 & 18/304 \\[6pt]
DBNILP   & 9/304 & 18/304 & 27/304  \\ \hline
\end{tabular}
\begin{center}
(b) Job-Shop
\end{center}
\end{minipage}
\caption{Fraction of near-optimal instances ($F(\textbf{x}) \leq \theta^*)$
    generated on each iteration of EODS.  In effect, this is
    an estimate of the recall (true-positive rate, or sensitivity)
    when predicting $F(\textbf{x}) \leq \theta^*$. The fraction $a/b$
    denotes that $a$ instances of $b$ are generated. }
\label{fig:results2}
\end{figure}
\section{Conclusions and Future Work}
In this paper we demonstrate that DBNs can be used as efficient samplers for EDA style optimization approaches. We further look at combining the sampling and feature discovery power of Deep Belief Networks with the background knowledge discovered by an ILP engine, with a view towards optimization problems that entail some degree of domain information. The optimization is performed iteratively via an EDA mechanism and empirical results demonstrate the value of incorporating ILP based features into the DBN. In the future we intend to combine ILP based background rules with more sophisticated deep generative models proposed recently \cite{Gregor14, Gregor15} and look at incorporating the rules directly into the cost function as in \cite{hu16}.
\bibliographystyle{plain}
\bibliography{trefs}
\end{document}